\documentclass[twoside,11pt]{article}

\usepackage{mlhc}
\usepackage{amsmath,amsbsy,amsfonts,bm}
\usepackage{epstopdf}
\usepackage{caption}
\usepackage{subcaption}
\usepackage{tikz}
\usetikzlibrary{arrows,positioning}
\usepackage{times}

\newcommand{\vc}{\mathbf}
\DeclareMathOperator*{\argmax}{arg\,max}

\ShortHeadings{Input-Output Non-Linear Dynamical Systems applied to Physiological
Condition Monitoring}{Georgatzis K., Williams C.K.I., Hawthorne C.}

\title{Input-Output Non-Linear Dynamical Systems applied to Physiological
Condition Monitoring}
       \author{\name Konstantinos Georgatzis \email {k.georgatzis@sms.ed.ac.uk} \\
       \addr School of Informatics \\
       University of Edinburgh\\
       Edinburgh, United Kingdom
       \AND
       \name Christopher K.I.\ Williams  \email {ckiw@inf.ed.ac.uk} \\
       \addr School of Informatics \\
       University of Edinburgh\\
       Edinburgh, United Kingdom\\
       Alan Turing Institute\\
       London, United Kingdom
       \AND
       \name Christopher Hawthorne \email {cwhawthorne@doctors.org.uk} \\
       \addr Institute of Neurological Sciences \\
       Queen Elizabeth University Hospital\\
       Glasgow, United Kingdom
       }

\begin{document}

\maketitle
\begin{abstract} {
We present a non-linear dynamical system for modelling the effect of drug infusions on the vital signs of patients admitted in Intensive Care Units (ICUs). More specifically we are interested in modelling the effect of a widely used anaesthetic drug (Propofol) on a patient's monitored depth of anaesthesia and haemodynamics. We compare our approach with one from the Pharmacokinetics/Pharmacodynamics (PK/PD) literature and show that we can provide significant improvements in performance without requiring the incorporation of expert physiological knowledge in our system.}
\end{abstract}
\section{INTRODUCTION}
\label{sec:intro}

We are concerned with the problem of predicting the effect of infused drugs on the vital signs of a patient in intensive care. In the biomedical literature this problem is broken into pharmacokinetics (PK) and pharmacodynamics (PD). Pharmacokinetics concerns the absorption, distribution, metabolism, and elimination of drugs from the body, while pharmacodynamics addresses the biochemical and physiological effects of drugs on the body. PK models are typically expressed as sets of ordinary differential equations (ODEs), while PD models are typically nonlinear functions relating drug concentration to observed vital signs.

From a machine learning point of view, these models consist of a linear dynamical system with control inputs (the drug infusion rates), and a non-linear output model. The PK models are based on quite a number of assumptions that are arguably not highly accurate representations of what is going on in the body (see section \ref{sec:PKPD_model} for more details). Thus our contribution is to take a more ``data-driven'' approach to the problem, fitting an input-output non-linear dynamical system (IO-NLDS\footnote{Inspired by the IO-HMM of \citet{bengio-frasconi-94}.}) to predict the vital signs based on input drug infusion rates. A notable difference to the PK/PD approach is that the latent process is not constrained to be mapped to any physiologically interpretable quantity and the model is free to learn any latent representation that might better explain the observed data. Our results show clear improvements in performance over the PK/PD approach.
\section{MODEL DESCRIPTION}
\label{sec:model}
\subsection{PK/PD model}
\label{sec:PKPD_model}
We start by describing the standard PK approach which is called \textit{compartmental modelling}. Compartmental models are an abstraction used to describe the rate of change of a drug's concentration in a patient by accounting for the processes of absorption, distribution, metabolism and excretion of the drug in different parts (compartments) of the human body. This approach then involves building a system of ODEs that describe the evolution of drug concentrations at different compartments. The standard compartmental model for modelling the pharmacokinetical properties of the anaesthetic drug Propofol is comprised of three compartments and dates back to \citet{gepts1987disposition}. Based on that work, a model widely used in practice is the one introduced by \citet{marsh1991pharmacokinetic} which has been further improved upon by \citet{white2008use}. This line of work only addresses the PK aspect of the task. In order to quantify the effect of the drug on the observed vital signs, one needs to add an extra ``effect'' compartment for each observed vital sign, and link the concentration at the effect compartment with the observed physiology as done e.g. in \citet{bailey2005drug}. 

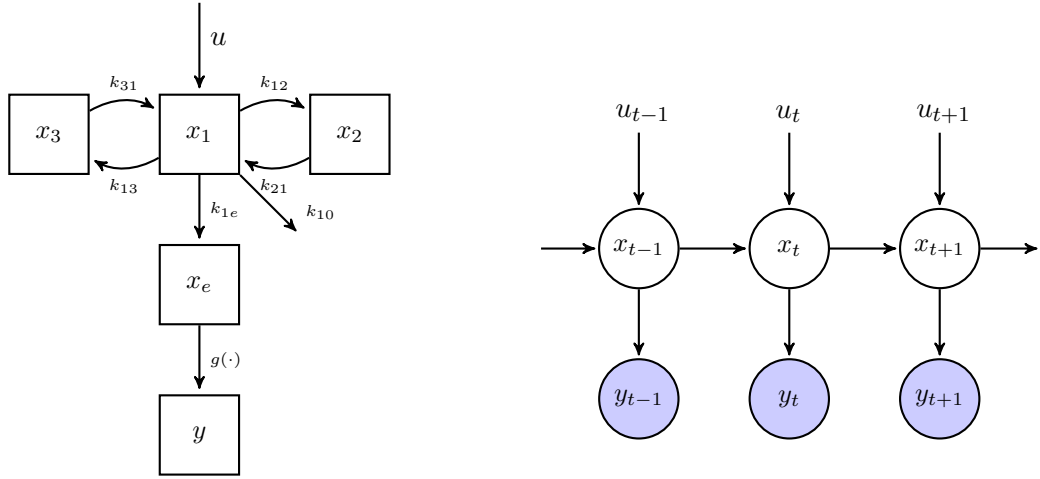
\begin{figure}[ht]
  \begin{subfigure}{.5\linewidth}
    \centering
    \begin{tikzpicture}[->,>=stealth',shorten >=2pt,auto,node distance=2cm,
    thick,main node/.style={rectangle,fill=white!20,minimum size = 30pt,draw,font=\fontsize{10}{1.5}\selectfont}]
  
    \node[main node] (1) {$x_{1}$};
    \node[main node] (2) [right of=1] {$x_{2}$};
    \node[main node] (3) [left of=1]  {$x_{3}$};
    \node[main node] (e) [below of=1]  {$x_{e}$};
    \node[main node] (y) [below of=e]  {$y$};
    \coordinate [below of=2] (h2);
    \coordinate [above of=1] (h0);
  
    \draw [->,>=stealth',shorten >=1cm] (1) to node {\tiny $k_{10}$} (h2);
    \draw [->,>=stealth',shorten <=0.25cm] (h0) to node {$u$} (1);
    \draw  (1) to node {\tiny $k_{1e}$} (e);
    \draw  (e) to node {\tiny $g(\cdot)$} (y);
  
    \path[every node/.style={font=\tiny} ]
  
    (1) edge [bend left] node {$k_{12}$} (2)
    (1) edge [bend left] node {$k_{13}$} (3)
    (2) edge [bend left] node {$k_{21}$} (1)
    (3) edge [bend left] node {$k_{31}$} (1);
    \end{tikzpicture}
  \end{subfigure}%
  ~
 \hspace{-2mm}
 \begin{subfigure}{.5\linewidth}
\centering
\begin{tikzpicture}[->,>=stealth',shorten >=1pt,auto,node distance=2cm,
  thick,state node/.style={circle,fill=white!20,minimum size = 30pt,draw,font=\fontsize{10}{1.5}\selectfont},
  obs node/.style={circle,fill=blue!20,minimum size = 30pt,draw,font=\fontsize{10}{1.5}\selectfont}]
  \node[state node] (1) {$x_{t-1}$};
  \coordinate [left of =1] (h1);
  \coordinate [above=2.5cm of 1] (hh1) ;
  \node[state node] (2) [right of=1] {$x_{t}$};
  \coordinate [above=2.5cm of 2] (hh2) ;
  \node[state node] (3) [right of=2] {$x_{t+1}$};
  \coordinate [above=2.5cm of 3] (hh3) ;
  \coordinate [right of=3] (h3);
  \node[obs node] (4) [below of=1] {$y_{t-1}$};
  \coordinate [left of =4] (h4);
  \node[obs node] (5) [below of=2] {$y_{t}$};
  \node[obs node] (6) [below of=3] {$y_{t+1}$};
  \coordinate [right of=6] (h6);
  
  \draw [->,shorten <=0.7cm] (h1) to node {} (1);
  \draw [->,shorten <=1.5cm] (hh1) to node {\hspace{-5.5mm}  $u_{t-1}$} (1);
  \draw [->,shorten <=1.5cm] (hh2) to node {\hspace{-4.25mm}  $u_{t}$} (2);
  \draw [->,shorten <=1.5cm] (hh3) to node {\hspace{-5.5mm}  $u_{t+1}$} (3);
  \draw [->,shorten >=0.7cm] (3)  to node {} (h3);
  
  \path[every node/.style={font=\fontsize{1}{1.5}\selectfont}]
    (1) edge node [right] {} (2)
    (1) edge      [below] node {} (4)
    (2) edge node [right] {} (3)
    (2) edge      [below] node {} (5)
    (3) edge      [below] node {} (6);
\end{tikzpicture}
\end{subfigure}
\caption{Left: a three compartment model for Propofol with one added effect site compartment. See text for details. Right: graphical model of IO-NLDS. The latent physiological state of a patient and the drug infusion rates at time $t$ are denoted by $\vc{x}_t$ and $\vc{u}_t$ respectively. The shaded nodes correspond to the observed physiological values, $\vc{y}_t$.}
\label{fig:PK-IO-NLDS}
\end{figure}

A graphical representation of this overall PK/PD approach is shown in Figure \ref{fig:PK-IO-NLDS} (left), where $x_{i}$ is the concentration of drug in compartment $i$ and $k_{ij}$\footnote{Parameter $k_{1e}$ is known as $k_{e0}$ in the PK literature.} is the drug's transfer rate from compartment $i$ to $j$. Thus, $x_{1}$ denotes the drug concentration in the central compartment, which is the site for drug administration and includes the intravascular blood volume and highly perfused\footnote{High ratio of blood flow to weight.} organs (e.g.\ heart, brain). The highest fraction of the administrated drug is assumed to reside in the central compartment. The remainder is diffused in two peripheral compartments which represent the body's muscle and fat. The drug's concentration in these two compartments is denoted as $x_{2}$ and $x_{3}$ respectively, while $x_{e}$ refers to the drug concentration at the effect site. For example, if one needs to measure the effect of an anaesthetic drug on a patient's consciousness level, then $x_{e}$ would refer to the patient's brain. Also, $u$ denotes the drug infusion rate and $k_{10}$ denotes the drug's elimination rate from the central compartment. Finally, one needs to establish the functional relationship between $x_{e}$ and the observed vital sign $y$. Since this relationship has a clear sigmoid shape, it is traditionally modelled via a generalised logistic function (also known as Richards' curve; see \citealp{richards1959flexible}) of the form: $g(x_{e}) = m + {(M-m)} / {(1+e^{-\gamma x_{e}})^{(1/\nu)}}$, where the parameters $m$, $M$ govern the lower/upper asymptote respectively, $\gamma$ controls the decrease rate and $\nu$ determines near to which asymptote maximum decrease occurs. This model can be described by a system of linear ODEs with the addition of a generalised logistic function as follows:
\begin{align}
\label{eq:3-CM_ODE}
\nonumber {dx_{1t}}/{dt}  &= -(k_{10} + k_{12} + k_{13})x_{1t} + k_{21}x_{2t} + k_{31}x_{3t} + u_{t} \ , \\
\nonumber {dx_{2t}}/{dt}  &=  k_{12}x_{1t} - k_{21}x_{2t} \ , \\
\nonumber {dx_{3t}}/{dt}  &=  k_{13}x_{1t} - k_{31}x_{3t} \ , \\
\nonumber {dx_{et}}/{dt}  &=  k_{1e}(x_{1t} - x_{et})\ , \\
                   y_{t}  &=  g(x_{et})\ .
\end{align}
An important aspect of this model is that the parameters associated with the PK part (i.e. the set of $k_{ij}$s) are considered fixed and their values have been determined by \citet{marsh1991pharmacokinetic} based on principles of human physiology. In \citet{white2008use} it was established that this PK model could be improved by allowing $k_{10}$ to vary according to a patient's age and gender. Hence, in this work, we follow the improved version of \citet{white2008use}.  
\vspace{-0.4cm}
\vspace{-0.4cm}
\subsection{IO-NLDS}
\label{sec:IO-NLDS}
In contrast to the PK/PD model we adopt an approach which makes no explicit use of expert physiological knowledge and is not restricted in associating the model's parameters with physiologically relevant quantities. Instead, the minimal assumptions made are that a latent temporal process with linear dynamics driven by control inputs (drug infusion), gives rise to observed quantities (vital signs) which are non-linearly dependent on the latent process. This gives rise to an input-output non-linear dynamical system (IO-NLDS) which is defined by the following joint distribution:
\begin{align}
\label{eq:model}
p(\vc{x},\vc{y}|\vc{u}) &= \hspace{1mm} p(\vc{x}_1|\vc{u}_1)p(\vc{y}_1|\vc{x}_1)\prod_{t=2}^Tp(\vc{x}_t|\vc{x}_{t-1},\vc{u}_t)p(\vc{y}_t|\vc{x}_t) \hspace{2mm} \ ,
\end{align}
where $\vc{x}_{t} \in R^{d_{x}}$, $\vc{u}_{t} \in R^{d_{u}}$ and $\vc{y}_{t} \in R^{d_{y}}$ denote the latent states, control inputs and observed vital signs at time $t$, and $T$ denotes the total length of the observed vital signs. We further assume that the random variables are distributed according to:
\begin{align}
\label{eq:udynamics}
 \mathbf{x}_{t} \sim \mathcal{N}\left(\mathbf{A}\mathbf{x}_{t-1} + \mathbf{B}\mathbf{u}_{t}, \mathbf{Q} \right),
\end{align}
\vspace{-1cm}
\begin{align}
\label{eq:uobservations}
 \mathbf{y}_{t} \sim \mathcal{N}\left(g(\mathbf{C}\mathbf{x}_{t}), \mathbf{R} \right) \ . 
\end{align}
The graphical model corresponding to these equations is shown in Figure \ref{fig:PK-IO-NLDS} (right). Equation (\ref{eq:uobservations}) encodes the assumption that the latent state is linearly projected onto the observation space via matrix $\vc{C}$, is subsequently non-linearly transformed via the generalised logistic function $g(\cdot)$, and corrupted by Gaussian observation noise with covariance $\vc{R}$ to give rise to the observations. Function $g(\cdot)$ is parameterised as in the PK/PD approach. The latent state itself is following linear dynamics, governed by matrix $\vc{A}$ and a linear transformation (via matrix $\vc{B}$) of control inputs, and additive Gaussian noise with covariance $\vc{Q}$. A notable difference to the PK/PD approach is that the latent process is not constrained to be mapped to any physiologically interpretable quantity and the model is free to learn any latent representation that might better explain the observed data. In contrast to the PK/PD model, the latent process exhibits a higher degree of flexibility, being unconstrained of any (simplifying) physiologically motivated assumptions and can thus be expected to model the patient-specific underlying dynamics in a more expressive way.
\subsection{PK/PD model as NLDS}
\label{sec:PK/PD_NLDS}
The PK/PD model as described in section \ref{sec:PKPD_model} does not incorporate any uncertainty while the IO-NLDS is a probabilistic model. In order to compare the two models, we cast the PK/PD model into the IO-NLDS form as described in eqs.\ (\ref{eq:udynamics}), (\ref{eq:uobservations}). This involves two steps: a) the discretisation of the continuous time dynamics as described by the system of ODEs in the first four equations of eq.\ (\ref{eq:3-CM_ODE}) and b) the addition of Gaussian noise on the discretised dynamics and on the non-linear output.  We can do this by setting the following parameters (assuming here $d_{y}=1$ for compactness) as $\vc{A}=\exp \{\vc{F}\Delta t\}, \vc{B}=[1 \ 0 \ 0 \ 0]^{\top}, \vc{C} = [0 \ 0 \ 0 \ 1]$, where $\vc{F}$ is described in the supplementary material.
The dynamics matrix $\vc{A}$ is the discretised version of its continuous time counterpart $\vc{F}$, which involves computing the matrix exponential of $\vc{F}$ times the discretisation step $\Delta t$ as described e.g.\ in \citet[sec.\ 5.3]{astrom2010feedback}. We note that the PK model provided by \citet{white2008use} provides estimates for the parameters involved in $\vc{F}$, except for $k_{1e}$. To fit an appropriate $k_{1e}$ we perform a fine-grained one-dimensional grid search per observed channel around a clinically relevant value. The noise matrices $\vc{Q}$ and $\vc{R}$ have the same interpretation as in the case of the IO-NLDS and can be learned in the same way, as described in section \ref{sec:learning}. Under this form, the PK/PD model can be seen as an IO-NLDS with a constrained parametric form, where the parameters $\vc{A}$, $\vc{B}$, and $\vc{C}$ are constrained in such a way as to capture the physiological processes involved with the infusion of an anaesthetic drug, as already described in section \ref{sec:PKPD_model}.
\subsection{Inference}
\label{sec:inference}
If the observation model was linear, then the overall model would be a linear dynamical system (LDS) and exact filtering would be feasible via the well-known Kalman filter \citep{kalman1960new}. With the addition of the nonlinear observation function $g(\cdot)$, exact inference becomes intractable and one must resort to some form of approximation. We adopt a sigma-point filtering approach, and more specifically we use the unscented Kalman filter (UKF), as described in \citet{sarkka2013bayesian}. The main component of UKF is the unscented transform (UT) as described in \citet{julier1996general} (although the term ``unscented'' was introduced later). The idea behind it is that a non-linear transformation of a Gaussian distribution can be approximated by first deterministically selecting a fixed number of points (called sigma-points) from that distribution to capture its mean and covariance, then computing the exact non-linear transformation of these points, and subsequently fitting a Gaussian distribution to the non-linearly transformed points. Under this procedure, the UKF reduces to applying the UT twice at each time step $t$: a) once to compute the predictive density $p(\vc{x}_{t}|\vc{y}_{1:t-1},\vc{u}_{1:t})$, by using the UT on the previous filtered density $p(\vc{x}_{t-1}|\vc{y}_{1:t-1},\vc{u}_{1:t-1})$ and b) to compute the current filtered density $p(\vc{x}_{t}|\vc{y}_{1:t},\vc{u}_{1:t})$ by using the UT on the predictive density to compute the current likelihood $p(\vc{y}_t|\vc{x}_t)$ and then using Bayes rule to obtain the required filtered (posterior) density. Since our model is governed by linear dynamics, the first step can be calculated by the standard Kalman filter equivalent step with both methods yielding the same results. 
\subsection{Learning}
\label{sec:learning}
The parameters of the proposed model that need to be learned are $\boldsymbol{\theta}=\{\boldsymbol{\mu}_1,\boldsymbol{\Sigma}_1,\vc{A},\vc{B},\vc{C},\vc{Q},\vc{R},\boldsymbol{\eta} \}$, where we assume that $\vc{x}_1 \sim \mathcal{N}(\boldsymbol{\mu}_1,\boldsymbol{\Sigma}_1)$ and $\boldsymbol{\eta} = \{\vc{m},\vc{M},\boldsymbol{\gamma},\boldsymbol{\nu}\}$ are the parameters of $g(\cdot)$. We learn maximum likelihood (ML) estimates of these parameters by using the expectation maximisation (EM) algorithm \citep{dempster1977maximum}. EM is an iterative algorithm that maximises the likelihood via maximising the following surrogate function:
\begin{align}
\label{eq:Q}
 \mathcal{Q}(\boldsymbol{\theta},\boldsymbol{\theta}^{old}) = \mathbb{E}_{p(\vc{x}|\vc{y},\vc{u},\boldsymbol{\theta}^{old})}[\log p(\vc{x},\vc{y}|\vc{u},\boldsymbol{\theta})] \ ,
\end{align}
which, due to the Markov properties of the model, can be further decomposed as:
\begin{align}
\label{eq:Q_long}
\nonumber \hspace{0mm} \mathcal{Q}(\boldsymbol{\theta},\boldsymbol{\theta}^{old}) &= \mathbb{E}_{p(\vc{x}_1|\vc{y}_{1:T},\vc{u}_{1:T},\boldsymbol{\theta}^{old})}[\log p(\vc{x}_1|\vc{u}_1,\boldsymbol{\theta})] + \mathbb{E}_{p(\vc{x}_t,\vc{x}_{t-1}|\vc{y}_{1:T},\vc{u}_{1:T},\boldsymbol{\theta}^{old})}[\log p(\vc{x}_t|\vc{x}_{t-1},\vc{u}_t,\boldsymbol{\theta})] \\
             &+ \mathbb{E}_{p(\vc{x}_t|\vc{y}_{1:T},\vc{u}_{1:T},\boldsymbol{\theta}^{old})}[\log p(\vc{y}_t|\vc{x}_{t},\boldsymbol{\theta})] \ .
\end{align}
These terms involve expectations with respect to the smoothing distributions $p(\vc{x}_t|\vc{y}_{1:T},\vc{u}_{1:T},\boldsymbol{\theta}^{old})$ and pairwise joint smoothing distributions $p(\vc{x}_t,\vc{x}_{t-1}|\vc{y}_{1:T},\vc{u}_{1:T},\boldsymbol{\theta}^{old})$. These distributions can be computed in general via the unscented Rauch-Tung-Striebel (URTS) smoother as described in \citet{sarkka2008unscented}. In our case, since the system's dynamics are linear, the backward smoothing step can be performed by the standard RTS smoother after obtaining the unscented filtered estimates during the forward filtering step via the UKF. Analogously to the UKF, where the UT is used to approximate the required expectations, the UT needs to be employed here to approximate the expectation appearing in the last term of the RHS of eq.\ (\ref{eq:Q_long}). More details are given in \citet{kokkala2014expectation}. 

Computing the required distributions and expectations in order to calculate eq.\ (\ref{eq:Q_long}) constitutes the E-step of EM. In the M-step the model parameters are set such that $\boldsymbol{\theta^{*}} \leftarrow \argmax_{\boldsymbol{\theta}} \mathcal{Q}(\boldsymbol{\theta},\boldsymbol{\theta}^{old})$. This maximisation step can be done analytically in the case of the LDS, but one needs to make use of numerical optimisation methods in the non-linear case. In our model, the subset of parameters $\boldsymbol{\theta^{*}}_{L} = \{\boldsymbol{\mu}_1,\boldsymbol{\Sigma}_1,\vc{A},\vc{B},\vc{Q}\}$ which correspond to the linear part of the model can be computed in closed form as shown in \citet{cheng2006modeling}, and $\vc{R}^{*}$ can be computed similarly to the linear case as shown in \citet[sec.\ 12.3.3]{sarkka2013bayesian}. We provide those estimates in the supplementary material. The remaining set of parameters $\boldsymbol{\theta}^{*}_{NL} = \{\vc{C},\boldsymbol{\eta}\}$ can be then optimised via numerical optimisation. In our experiments we use the Broyden-Fletcher-Goldfarb-Shanno (BFGS) algorithm (see e.g. \citealp{fletcher2013practical}). We note that EM is a natural choice in our case since our model involves a linear sub-component which can be exploited in the decomposition of eq.\ (\ref{eq:Q_long}) to derive a subset of parameters in closed form. In a fully non-linear case however one could use a numerical optimisation procedure to directly maximise the likelihood function instead of a surrogate function, as argued in \citet{kokkala2015sigma}. To the best of our knowledge, this is the first time that the UKF and URTS are used within EM for ML estimation of parameters that include also control inputs.

\subsection{Related work}
\label{sec:review}
The PK/PD model presented in section \ref{sec:PKPD_model} is a standard approach in the PK/PD literature that stems from a long line of research. Starting with the PK component, in \citet{benet1972general}, the Laplace transform method is presented as a way of analysing multi-compartmental PK models. Work conducted by \citet{gepts1987disposition} is focussed on the PK properties of the anaesthetic drug Propofol under a continuous drug infusion, as opposed to bolus injections which used to be standard practice. In that work, it was concluded that a three-compartmental PK model best fits the observed data, and physiologically relevant parameters for this model were established. In \citet{marsh1991pharmacokinetic}, the well-known (in PK literature) \textit{Marsh} model was presented, which investigated the previous model and suggested a new set of parameters that better reflected the distribution and elimination of Propofol so as to include children. In \citet{white2008use}, it was shown that the \textit{Marsh} model should be modified to allow for the elimination rate ($k_{10}$) of Propofol from the central compartment to vary depending on a patient's age and gender. 

The PD component has a similarly long history. In \citet{colburn1981simultaneous}, one, two and three compartmental PK models with the addition of an effect site compartment to model PD effects were investigated. Similarly, in \citet{fuseau1984simultaneous}, a three compartmental PK model with an additional effect site compartment was studied. An approach which is closer to our proposed PK/PD model is presented in \citet{bailey2005drug}, where a three-compartmental PK model with an effect site compartment is cast as a state space model and a sigmoid function links the drug concentration at the effect site to the patient's observed level of consciousness. However, that model is non-stochastic, and is evaluated on simulated data with parameters fixed on a priori known values.

In terms of research which is closer to the machine learning community, there has been work which tackles various tasks of interest with respect to physiological monitoring in ICUs. In \citet{quinn2009factorial}, a factorial switching linear dynamical system (FSLDS) was used to infer artifactual and physiological processes of interest, whereas in \citet{georgatzis2015discriminative} a discriminative SLDS is used for the same purposes. In \citet{lehman2015physiological}, a switching vector autoregressive process was used to extract features from vital signs which were used as input in a logistic regression classifier to predict patient outcome, while work by \citet{nemati2013learning} was focussed on discriminatively training a SLDS for learning dynamics associated with patient outcome. These methods are primarily concerned with the classification of events of interest ignoring any administered drugs. Additionally, all of these works use linear models as their core elements, while our model is non-linear. More similar work to ours is presented by \citet{enright2011clinical} and \citet{enright2013bayesian}, where a non-linear dynamical system is developed to model the glucose levels of patients under the intravenous administration of glucose and insulin. However, their approach relies solely on converting existing systems of ODEs into a probabilistic model, which in our case corresponds only to the PK/PD model and is clearly outperformed by the fully data-driven approach.

In terms of methodology, the task of inference in non-linear dynamical models has been thoroughly explored and methods such as the extended Kalman filter (see e.g. \citealp[sec.\ 5.2]{sarkka2013bayesian}), the UKF \citep{wan2000unscented} and the particle filter \citep{gordon1993novel} have been proposed. The UKF is shown empirically to outperform the extended KF (see e.g.\ \citealp[Ch.\ 7]{haykin2001kalman}), and was chosen over the particle filter (PF) because the PF can require ``orders of magnitude'' (see \citealp{wan2000unscented}) more sample points compared to the UKF's fixed, small number of sigma-points to achieve high accuracy, rendering it computationally prohibitive for a real-time application such as ours. Learning is not as thoroughly explored as inference in NLDSs but nonetheless there has been considerable work recently. In \citet{schon2011system} the EM algorithm is used in conjunction with PF for parameter estimation in NLDSs, while in \citet{gavsperin2011application} and \citet{kokkala2014expectation} the EM is used in conjunction with the UKF for the same task. However, they do not include control inputs in their formulation as we do.
\vspace{-2.5mm}
\section{EXPERIMENTS}
\label{sec:experiments}
In this section we describe experiments that were conducted in order to establish if the newly proposed approach can model accurately the effect of drug infusions on the observed physiology of patients in ICUs. To this end, we compare the IO-NLDS and the PK/PD model for the task of predicting the effect of Propofol on patients' vital signs.
\vspace{-2.5mm}
\subsection{Data Description}
\label{sec:data}
The dataset comprises of 40 Caucasian patients admitted in the neuro ICU of the Golden Jubilee National Hospital in Glasgow, Scotland. All patients were spontaneously breathing with the maximum airway intervention being an oropharyngeal airway, and none was classified higher than Class 2 with respect to illness severity according to the American Society of Anesthesiologists. Two controlled Propofol infusion protocols were investigated on these patients pre-operatively during a period of approximately 45 minutes. This investigation was part of an independent clinical study and the data have been anonymised. Each patient was randomly assigned to one of the two protocols. The first protocol involved a target Propofol concentration of 2 $\mu$g/ml for the first 15 minutes followed by a target concentration of 5 $\mu$g/ml for the next 15 minutes and 2 $\mu$g/ml for the last 15 minutes. The second protocol was the inverse of the first with a 5-2-5 $\mu$g/ml target sequence. The drug pumps which automatically administer Propofol calculate internally the desired infusion rates and these rates were used as control inputs in our model. Propofol doses are at peak during the early phase of the protocol making it very likely that the maximum haemodynamic effects will be seen during the study period and not after discontinuation of the drug. Also, during the study very few interventions were required in the case of hypotension (treated by small incremental doses of either ephedrine or Metaraminol) and bradycardia (treated by Glycopyrolate). The frequency and duration of these interventions were deemed sufficiently low so as to treat the overall dataset as unaffected by them. During the duration of the protocol, six vital signs were recorded, namely systolic, mean and diastolic blood pressure (BPsys, BPmean, BPdia), heart rate (HR), respiratory rate (RR) and the bispectral index (BIS). BIS was recorded only in 27 of the 40 patients. Of those channels, the first five were measured at a time interval $\Delta t = 15$ seconds and BIS was measured every 5 seconds and was subsequently subsampled to 15 seconds. BIS is a standard index (see e.g. \citealp{rampil1998primer}) that is based on features extracted from the spectrum of a patient's electroencephalogram (EEG) and measures depth of anaesthesia, with values ranging between $0-100$ (lower values representing deeper anaesthesia). From these values, BPsys, BPdia and BIS were expected to be affected by Propofol administration and the signals were of adequate quality so that they could be further analysed.
\vspace{-2.5mm}
\subsection{Model fitting}
\label{sec:model_fit}
For both models we use EM as described in section \ref{sec:learning} to learn the parameters on each patient separately and then obtain predicted values using those fitted parameters. For both models, 100 iterations of EM where used and the BFGS algorithm was run with 1000 function evaluations during the first 10 iterations and 100 evaluations during the remaining iterations. An important advantage of our model's linear dynamics is that we can enforce stability constraints which are of paramount importance in a real-world setting. If the systems dynamics matrix, $\vc{A}$, becomes unstable (i.e if the modulus of its largest eigenvalue is greater than one) we project this matrix back to the space of stable matrices such that it is also closer (in a least-squares sense) to the originally learned matrix. We follow the approach proposed by \citet{siddiqi2007constraint} to achieve this, which involves solving for a quadratic program inside EM. This process is very fast and is performed only if an instability is observed. We also constrain the noise covariance matrices $\vc{Q}$ and $\vc{R}$ to be diagonal. Furthermore, we set $d_{x}=4$ for the IO-NLDS. We fix it to this value because it directly mirrors the four compartments of the PK/PD model, and thus allows for a more direct and fair comparison between the two models. Under these assumptions, the IO-NLDS has $65$ parameters compared to $50$ in the case of the PK/PD model. The higher number of parameters in the case of the IO-NLDS reflects its greater expressive power since it is not constrained by physiological assumptions. We note, however, that in the general case the IO-NLDS' latent space dimensionality does not have to be restricted in such a way. In contrast with the PK/PD model, one has the flexibility to decide on alternative latent space dimensionalities; e.g.\ by making use of an information criterion to determine $d_{x}$ under a more formal measure of optimality, turning the whole process into a standard model order selection procedure. In the PK/PD case, a higher latent space dimension would correspond to an increased number of compartments, which could require years of additional research to validate physiologically appropriate extensions of the already existing literature. However, to perform a more fair model comparison between the two models, we also compute Bayesian Information Criterion (BIC) scores to account for the increased number of parameters in the case of the IO-NLDS. Finally, the UT involves determining three parameters $\alpha$, $\beta$, $\kappa$ (see supplementary material for details). Following \citet[sec.\ 5.5]{sarkka2013bayesian}, we set $\alpha=1$, $\beta=0$, and $\kappa=3-d_{x}$.
\subsection{Results}
\label{sec:results}
We use the standardised mean squared error (SMSE)\footnote{$SMSE = MSE/var_{y}$, where $MSE$ is the mean squared error and $var_{y}$ is the variance of the observed vital sign.} between predictions and actual observations as our evaluation metric, and also provide representative examples of curve fits on the observed data produced by the two models. The SMSE takes into account the variance of the observed data and provides a natural baseline with SMSE $=1$ denoting the mean prediction. For both models we compare the SMSE between model predictions and measured outputs of BPsys, BPdia and BIS across a number of different prediction horizons; namely $h=$ 1, 10 and 20-step ahead predictions, (corresponding to 15 seconds, 2.5 minutes and 5 minutes respectively). We also evaluate the SMSE for the ``free-running'' case, which corresponds to predictions for the whole duration of the protocol. These time intervals, apart from the 1-step interval, were decided as clinically relevant. We include results on the 1-step ahead prediction task, since this is a standard evaluation task when assessing predictive performance of dynamical models. 

\begin{figure}[]
        \hspace{-5cm}
        \begin{center}
        \begin{subfigure}[ht]{0.333\textwidth}
                \hspace{-0.4cm}
                \centering
                \includegraphics[width=1\textwidth, height=4.5cm]{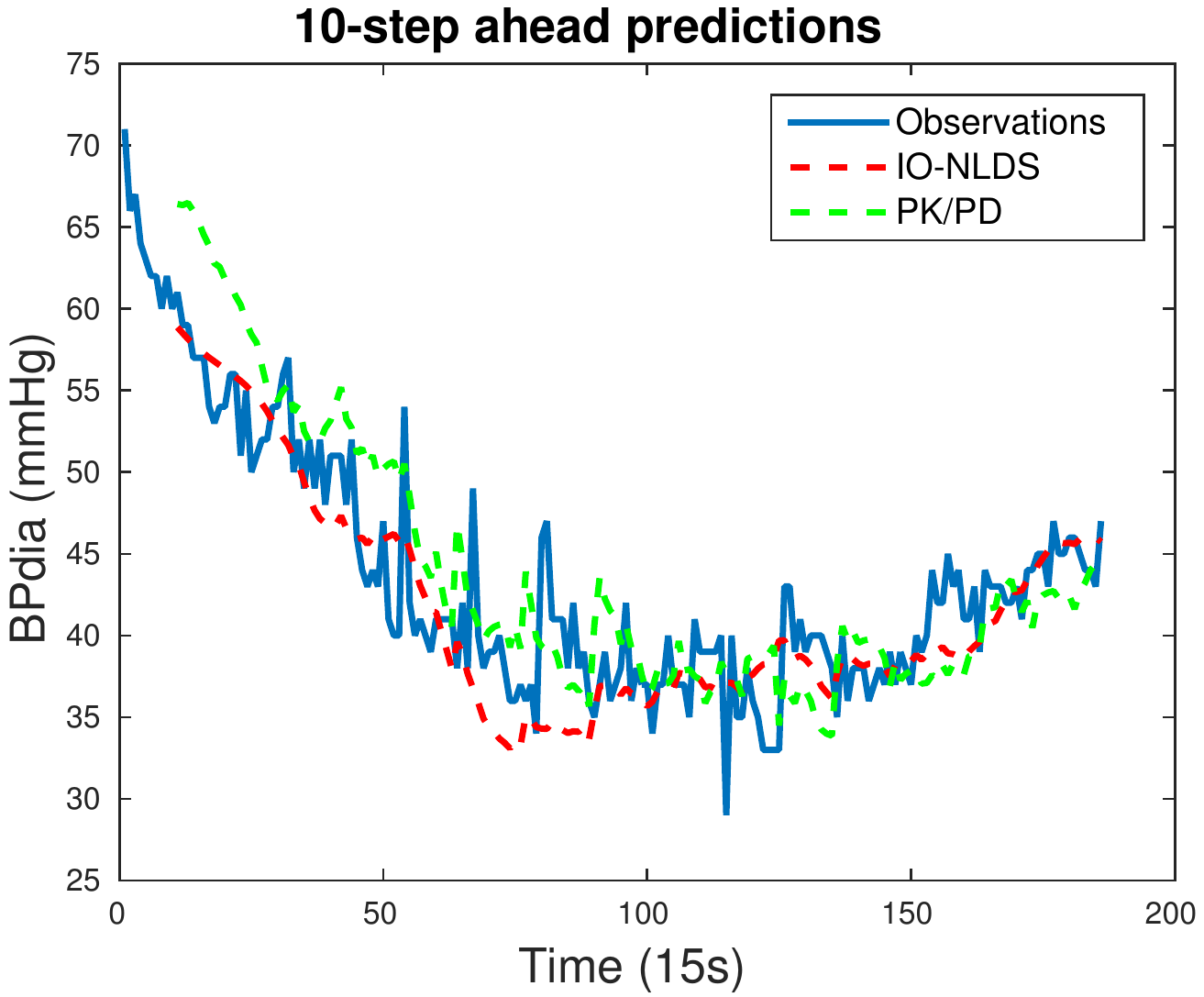}
        \end{subfigure}%
        ~
        \begin{subfigure}[ht]{0.333\textwidth}
                \hspace{-0.4cm}
                \centering
                \includegraphics[width=1\textwidth, height=4.5cm]{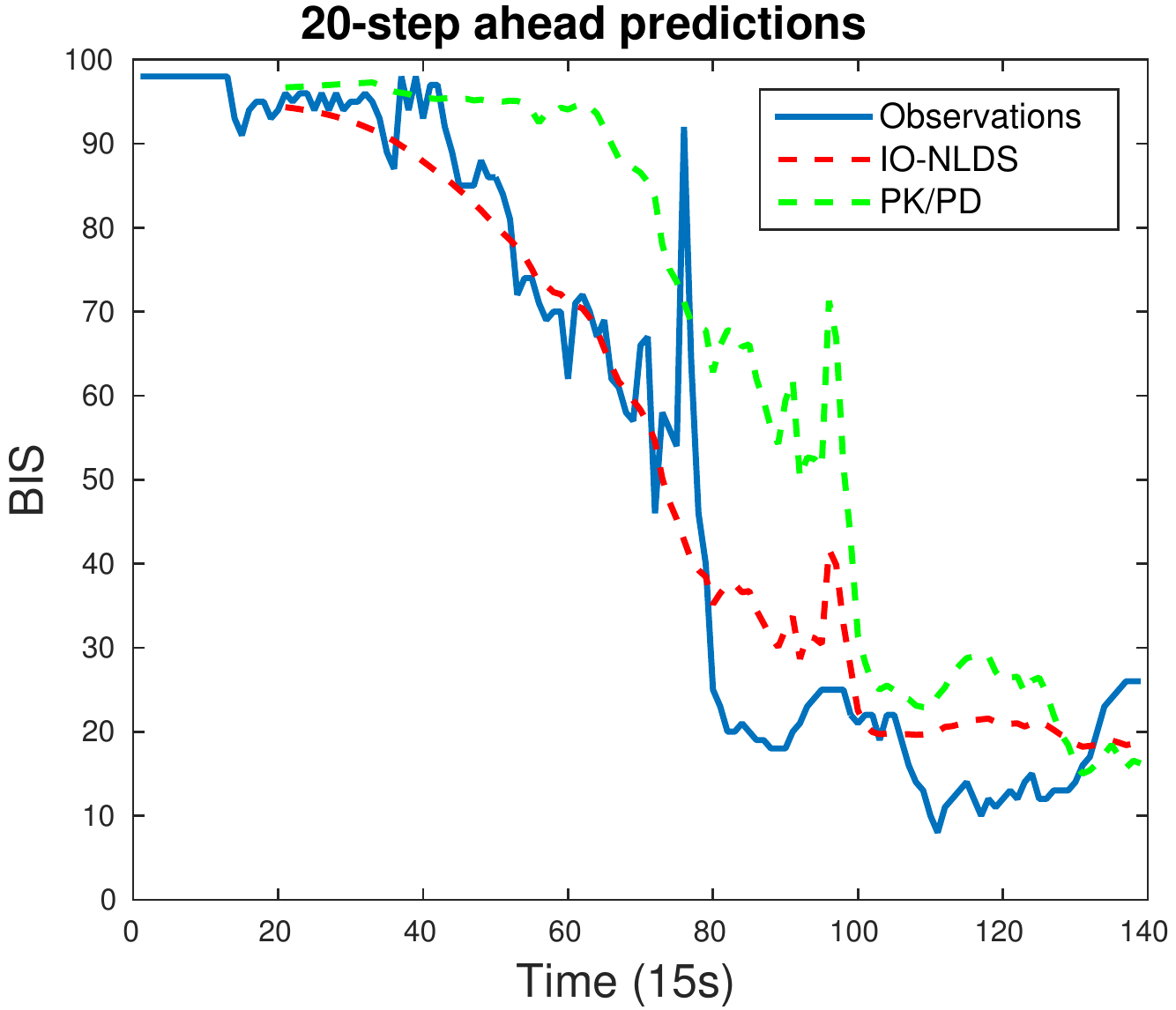}
        \end{subfigure}%
        ~
        \begin{subfigure}[hb]{0.333\textwidth}
                \hspace{-0.4cm}
                \centering
                \includegraphics[width=1\textwidth, height=4.5cm]{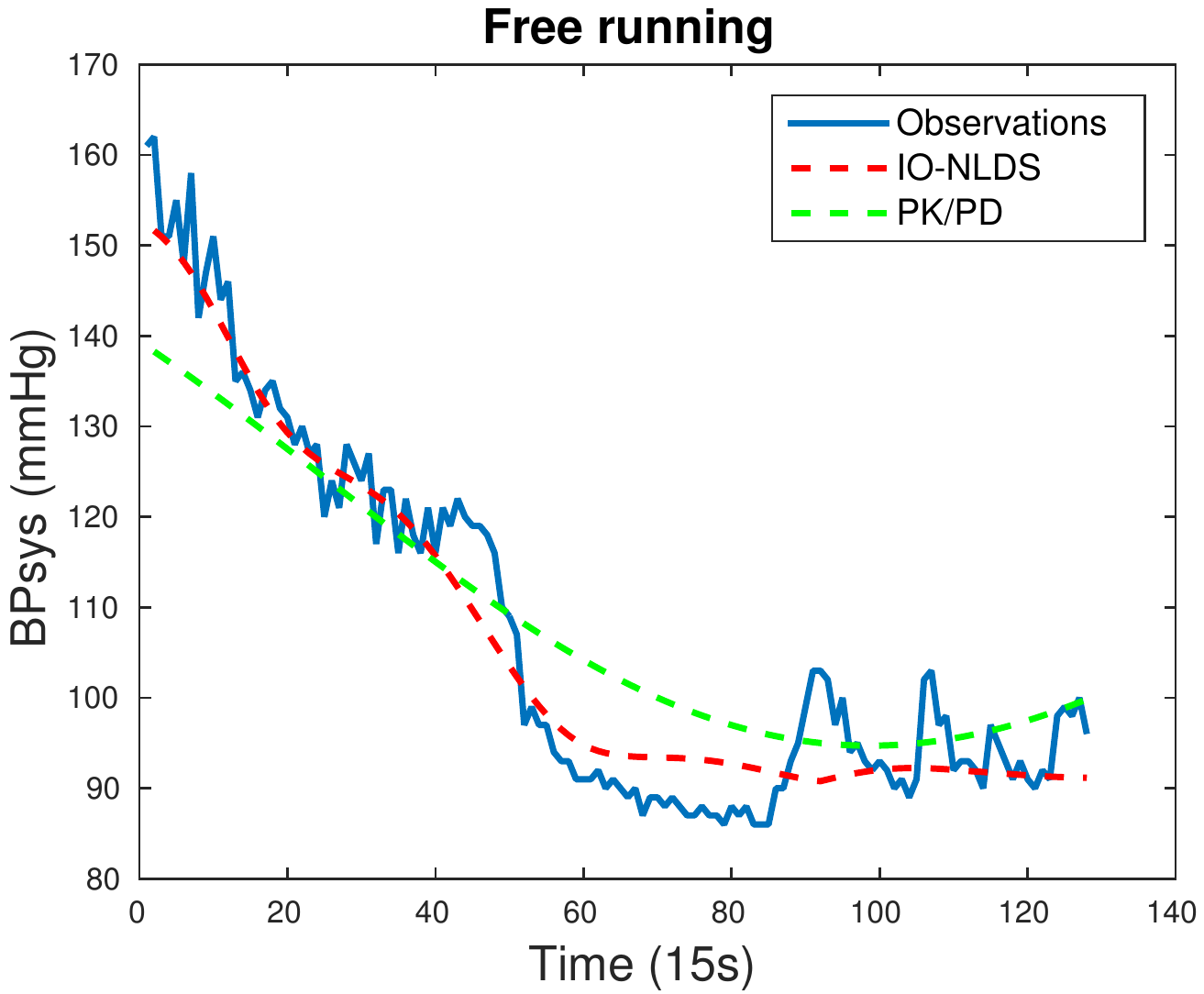}
        \end{subfigure}%
        \caption{Examples of observed vital sign traces and the predictions made by the IO-NLDS (red dashed line) and the PK/PD model (green dashed line) for different prediction horizons.}
        \label{fig:traces}
        \end{center}
\vspace{-0.75cm}
\end{figure}

In Figure \ref{fig:traces}, three representative examples of observed vital signs and the fitted traces produced by the two models are presented. In the left panel, an example on the 10-step ahead prediction task for the BPdia channel is given. Both models manage to capture the observed temporal structure with satisfactory accuracy. In the middle panel, an example of the 20-step ahead task on BIS is given. Here the difference in performance is clearer, with the IO-NLDS tracking the temporal evolution of the observed signal more accurately, especially during the time that BIS decreases more rapidly which corresponds to a rapid increase in the patient's depth of anaesthesia. Finally, in the right panel, the ``free-running'' predictions of the two models are shown along with the observed BPsys channel. The IO-NLDS manages to stay very close to the actual signal, while on the other hand the PK/PD model showcases a considerably higher degree of error in its predictions, as it has not managed to learn an appropriate rate of decrease on its predictions. In all cases, the IO-NLDS manages to stay closer to the observed signal especially at the beginning of the protocol during the steepest decrease of the observed signal. This constitutes the most critical phase of the protocol during which the uncertainty of an anaesthetist about the patient's reaction to the drug is considerable and the risk of an undesirable episode (e.g.\ hypotension) is at its highest. Therefore, accurate predictions during that phase are much more critical compared to the rest of the signal.

\begin{figure}[ht]
\vspace{-0.75cm}
        \begin{center}
        \begin{subfigure}[ht]{0.333\textwidth}
                \centering
                \includegraphics[width=1\textwidth, height=4.5cm]{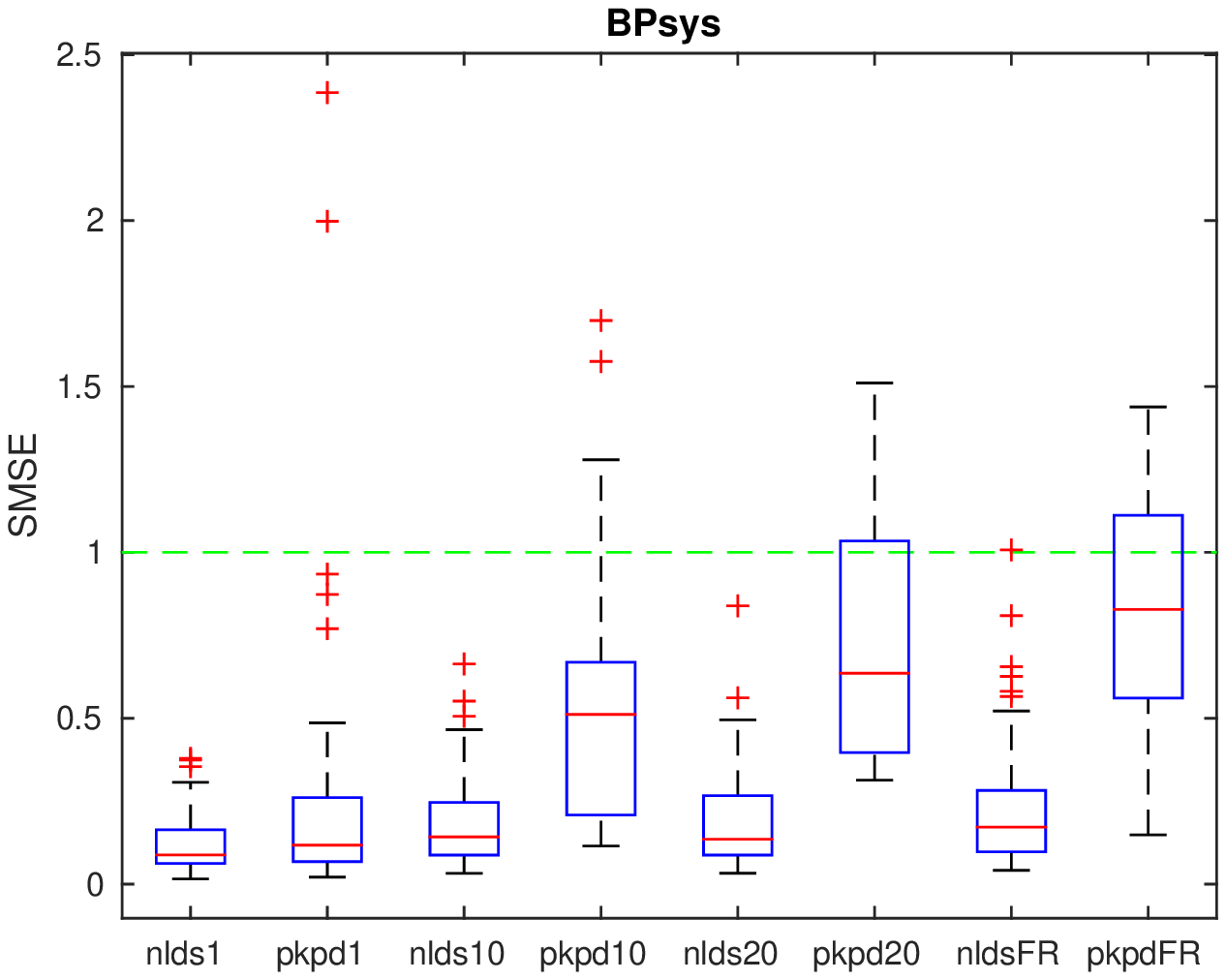}
        \end{subfigure}%
        ~
        \begin{subfigure}[ht]{0.333\textwidth}
                \centering
                \includegraphics[width=1\textwidth, height=4.5cm]{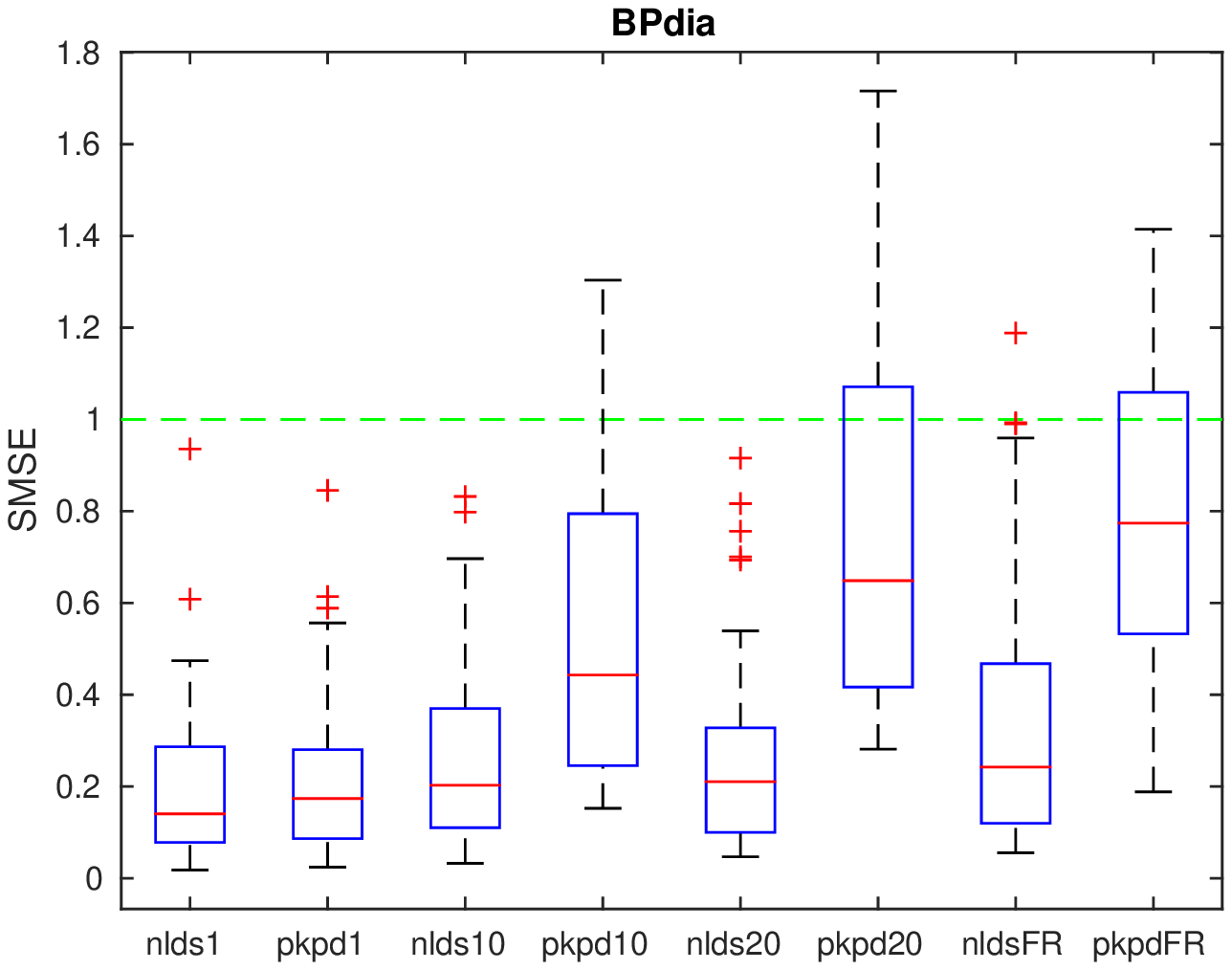}
        \end{subfigure}%
        ~
        \begin{subfigure}[hb]{0.333\textwidth}
                \centering
                \includegraphics[width=1\textwidth, height=4.5cm]{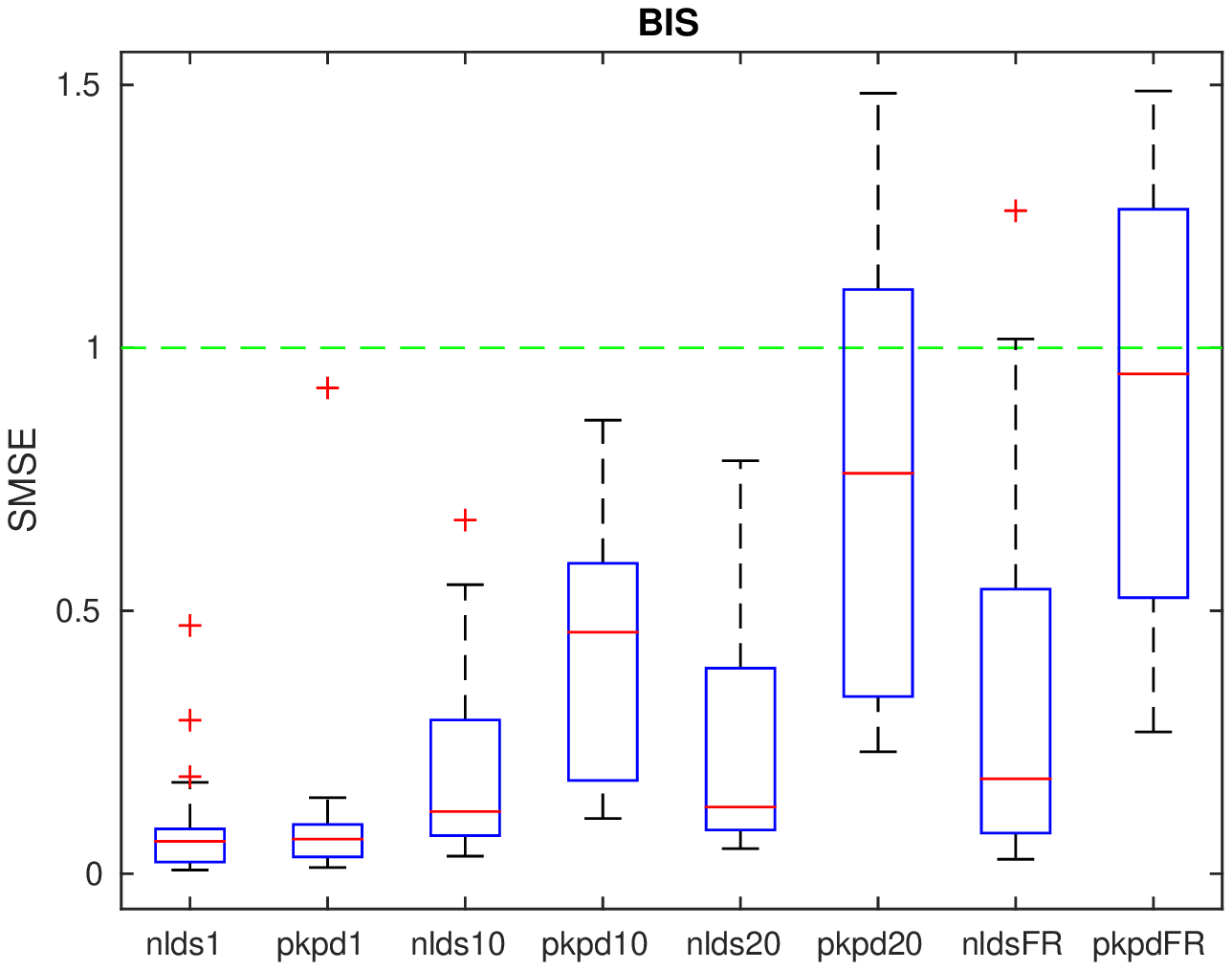}
        \end{subfigure}%
        \caption{Summarised SMSE results for IO-NLDS compared to PK/PD for BPsys, BPdia and BIS across the four prediction horizons for all patients. The green dashed line corresponds to the mean predictions. The x-axis is labelled according to the model name followed by the prediction horizon.}
        \label{fig:boxplots}
        \end{center}
\vspace{-0.75cm}
\end{figure}

The results for the three channels and the four prediction horizons are shown as boxplots in Figure \ref{fig:boxplots}, where the central mark denotes the median, the edges of the box are the lower and upper quartiles and the whiskers extend to outliers with extreme outliers being denoted separately by a red cross. For all three channels, the IO-NLDS' predictions are consistently more accurate than the PK/PD model's across all four prediction horizons. Moreover, the difference in favour of the IO-NLDS becomes more obvious as the prediction horizon increases. Also, in almost all cases the predictive errors for both models are increased as the prediction horizon increases, as expected.

A summary of the mean SMSEs of the two models per prediction horizon and per channel is shown in Table \ref{tab:meanSMSE}. The IO-NLDS' prediction errors are consistently lower than the PK/PD model's and by a large margin in most cases. Both models' errors are low at the 1-step ahead prediction level. In the same table, we also provide BIC scores per prediction horizon for both models to account for IO-NLDS' higher number of parameters. The BIC is defined as: $BIC = -2 \ln(L) + b \ln(N)$, where $L$ is the data likelihood under the model, $b$ is the number of free parameters and $N$ is the number of data points. The IO-NLDS achieves a lower (better) score in all cases and thus demonstrates that the increase in likelihood is not just an effect of higher statistical capacity but rather that the IO-NLDS manages to model the observed temporal structure better than the PK/PD model.

A more fine-grained comparison shows that in $92\%$ of all investigated cases the accuracy of the IO-NLDS predictions is better than the PK/PD model's (see Figure S1 in the supplementary material). Finally, we perform twelve paired, right-tailed t-tests (one per channel and per prediction horizon) on the differences of the obtained SMSEs  $(SMSE_{PKPD}-SMSE_{NLDS})$ across the 40 patients. The null hypothesis that the mean of this difference is zero, is rejected in $10$ out of $12$ cases with $p$-values ranging from $1.6\times 10^{-2}$ to $4.8\times 10^{-17}$. In the case of the 1-step ahead predictions on BPdia and BIS, $p$-values of $0.14$ and $0.37$ respectively were obtained and thus the null hypothesis could not be rejected.
\vspace{-0cm}
\begin{table*}[h]
\setlength{\tabcolsep}{2.75pt}
\caption{Comparison of IO-NLDS and PK/PD model with respect to mean SMSE per channel and prediction horizon and mean BIC per prediction horizon. Lower numbers are better.}
\vspace{-0.5cm}
\label{tab:meanSMSE}
\begin{center}
\resizebox{\textwidth}{!}{%
\begin{tabular}{|c | l l l | l | l l l | l | l l l | l | l l l | l |}
\hline
\multicolumn{1}{|c}{\bf SMSE/BIC}   &\multicolumn{4}{|c}{1-step} &\multicolumn{4}{|c}{10-step} &\multicolumn{4}{|c}{20-step} &\multicolumn{4}{|c|}{free-running}\\
\cline{2-17} 
         & {BPs} & {BPd} & {BIS} & {BIC}   & {BPs} & {BPd} & {BIS} & {BIC}   & {BPs} & {BPd} & {BIS} & {BIC}   & {BPs} & {BPd} & {BIS} & {BIC} \\
\hline 
IO-NLDS  &$0.13$ &$0.20$ &$0.08$ &$2592$   &$0.19$ &$0.27$ &$0.21$ &$4458$   &$0.20$ &$0.27$ &$0.25$ &$6656$   &$0.25$ &$0.36$ &$0.33$ &$16588$\\      
PK/PD    &$0.30$ &$0.23$ &$0.10$ &$2709$   &$0.52$ &$0.53$ &$0.41$ &$5314$   &$0.73$ &$0.75$ &$0.74$ &$7919$   &$0.84$ &$0.82$ &$0.89$ &$19008$\\
\hline
\end{tabular}}
\end{center}
\end{table*}
\vspace{-0.5cm}
\section{DISCUSSION}
\label{sec:discussion}
We first describe a use case scenario for the IO-NLDS and then provide suggestions for future work.

The current practice in an ICU is that an anaesthetist will start drug administration on a patient with an initial target dosage and then readjust it according to the evolution of the patient's vital signs. This implies that the anaesthetist needs to be at bedside continuously and for the whole duration of the drug administration, and by definition can only be reactive to potential vital sign deteriorations. Also, this monitoring is not always straightforward. For example, during drug induction when the risk of hypotension is high, the anaesthetist's attention is divided as they will be also managing the airway (e.g.\ by inserting an endotracheal tube). Therefore a use case for the system is to take a clinician-specified drug infusion protocol, and make predictions for the time-course of the vital signs. The clinician could then inspect these forecasts, and either continue with the original protocol or adjust it accordingly. (In that case the model could also automatically notify the clinician, e.g.\ if a threshold is crossed by the predicted values). This extra functionality would help the anaesthetist to be proactive and to avert undesirable episodes for the patient (e.g.\ an episode of hypotension). The PK/PD literature has become very focussed on compartmental models, but more data-driven models may give rise to better predictions which could thus translate to better clinical outcomes.

We have presented a purely data-driven approach for the important application of modelling drug effects on the physiology of patients in ICUs. We show that our new approach outperforms the previous expert-knowledge based one. To the best of our knowledge, this is the first time that a complex, real-world model has been fully learned via a combination of unscented filters and EM, making this a promising direction for related tasks. As our model is drug-agnostic, we plan to apply it to other drugs in future work. Finally, in our experiments a separate NLDS model was fitted per patient. However, the model would be more powerful if its parameters could be predicted from patient covariates (e.g.\ age, gender etc.\ ) and this is the target of our future work.
\subsubsection*{Acknowledgements}
We extend our thanks to Martin Shaw for preprocessing code and valuable discussions. Author KG was funded by the Scottish Informatics and Computer Science Alliance. The work of CW is supported in part by EPSRC grant EP/N510129/1.
\bibliographystyle{natbib}
\bibliography{ionlds.bib}

\appendix
\section*{SUPPLEMENTARY MATERIAL}
We provide details about the PK/PD model's continuous time dynamics matrix $\vc{F}$, the UT and the EM algorithm.

\subsection*{S.1\space\space PK/PD continuous time dynamics matrix}
\label{sec:PK_PD_F}
\begin{align}
\nonumber \vc{F} &= \begin{bmatrix} 
                     -(k_{10} + k_{12} + k_{13})   &k_{21}      &k_{31}  &0       \\ 
                     k_{12}                        &-k_{21}     &0       &0       \\
                     k_{13}                        &0           &-k_{31} &0       \\
                     k_{1e}                        &0           &0       &-k_{1e} \\
          \end{bmatrix}\ .
\end{align}

\subsection*{S.2\space\space Unscented transform}
\label{sec:app_UT}
The UKF and EM involve the calculation of integrals of the form $\mathbb{E}_{p(\vc{x})}[g(\vc{x})] = \int g(\vc{x})\mathcal{N}(\vc{x}|\boldsymbol{\mu},\boldmath{\Sigma}) d\vc{x}$. Since $g(\cdot)$ is non-linear, these integrals are approximated via Gaussian cubature (i.e. multidimensional Gaussian quadrature) as: $\mathbb{E}_{p(\vc{x})}[g(\vc{x})] \approx \sum_{i}w_{i}g(\vc{x}_i)$. Following similar notation to \citet[sec.\ 18.5.2.1]{murphy2012machine}, we first define a set of $2d+1$ sigma points $\vc{x}_i$:
\begin{align}
\nonumber \vc{x} = \{\boldsymbol{\mu},\{\boldsymbol{\mu} + (\sqrt{(d+\lambda)\boldsymbol{\Sigma}})_{i}\},\{\boldsymbol{\mu} - (\sqrt{(d+\lambda)\boldsymbol{\Sigma}})_{i}\} \} \ ,
\end{align}
where $\boldsymbol{\Sigma}_{i}$ denotes the $i$'th column of matrix $\boldsymbol{\Sigma}$ and $i=1,...,d$. This set of sigma points is propagated via $g(\cdot)$, producing a new set of transformed points $\vc{y}_{i}$ with mean and covariance:
\begin{align}
\nonumber \boldsymbol{\mu}_{y}    &= \sum_{i=0}^{2d} w_{m}^{i}\vc{y}_{i} \ , \\
\nonumber \boldsymbol{\Sigma}_{y} &= \sum_{i=0}^{2d} w_{c}^{i}(\vc{y}_{i}-\boldsymbol{\mu}_{y})(\vc{y}_{i}-\boldsymbol{\mu}_{y})^{\top} \ , 
\end{align}
with weights $w$'s defined as:
\hspace{-1cm}
\begin{align}
\nonumber w_{m}^{0} &= \dfrac{d}{d+\lambda} \ , \\
\nonumber w_{c}^{0} &= \dfrac{d}{d+\lambda} + (1 - \alpha^2 + \beta) \ , \\
\nonumber w_{m}^{i}, w_{c}^{i} &= \dfrac{1}{2(d+\lambda)} \ ,
\end{align}
where $\lambda = \alpha^2(d+\kappa) - d$ and $\alpha$, $\beta$, and $\kappa$ are method-specific parameters. 
\subsection*{S.3\space\space EM}
\label{sec:app_EM}
We provide ML estimates for the parameters involved in eq.\ (\ref{eq:Q_long}) for the $i$'th iteration of EM that can be derived in closed form. We first define the following:
\begin{align}
\nonumber S_{x-x-} &= \sum_{t=2}^{T} \vc{P}_{t-1|T} + \boldsymbol{\mu}_{t-1|T}\boldsymbol{\mu}_{t-1|T}^{\top} \ , \\
\nonumber S_{xx-} &= \sum_{t=2}^{T} \vc{P}_{t,t-1|T} + \boldsymbol{\mu}_{t|T}\boldsymbol{\mu}_{t-1|T}^{\top} \ , \\
\nonumber S_{xx} &= \sum_{t=2}^{T} \vc{P}_{t|T} + \boldsymbol{\mu}_{t|T}\boldsymbol{\mu}_{t|T}^{\top} \ , \\
\nonumber S_{x-u} &= \sum_{t=2}^{T}  \boldsymbol{\mu}_{t-|T}\boldsymbol{u}_{t}^{\top} \ , \\
\nonumber S_{xu} &= \sum_{t=2}^{T}  \boldsymbol{\mu}_{t|T}\boldsymbol{u}_{t}^{\top} \ , \\
\nonumber S_{uu} &= \sum_{t=2}^{T}  \boldsymbol{u}_{t}\boldsymbol{u}_{t}^{\top} \ , 
\end{align}
where $\boldsymbol{\mu}_{t|T}$, $\vc{P}_{t|T}$ denote the smoothed mean and covariance estimates at time $t$ and $\vc{P}_{t,t-1|T}$ denotes the smoothed pairwise covariance estimates at times $t-1,t$, as outputted by the URTS smoother. We then have:
\begin{align*}
  \boldsymbol{\mu}_{1}^{i}    &= \boldsymbol{\mu}_{1|T}  \ ,  \\
  \boldsymbol{\Sigma}_{1}^{i} &= \vc{P}_{1|T} \ ,  \\
  [\vc{A}^{i} \ \vc{B}^{i}]   &= [S_{xx-} \ S_{xu}] \begin{bmatrix}   
                                                                           S_{x-x-}  &S_{x-u}   \\ 
                                                                           S_{x-u}^{\top} &S_{uu}        
                                                             \end{bmatrix}^{-1}  \ ,  \\ 
 \vc{Q}^{i} &= \dfrac{1}{T-1} (S_{xx} - \vc{A}^{i}S_{xx-}^{\top} - \vc{B}^{i}S_{xu}^{\top})  \ . 
\end{align*}
Finally, we have:
\begin{flalign*}
 \vc{R}^{i} = \dfrac{1}{T} \sum_{t=2}^{T} \mathbb{E}_{p(\vc{x}_t|\vc{y}_{1:T},\vc{u}_{1:T},\boldsymbol{\theta}^{i-1})} [(\vc{y}_t - g(\vc{x}_t,\boldsymbol{\theta}_{NL})) (\vc{y}_t - g(\vc{x}_t,\boldsymbol{\theta}_{NL}))^{\top}] \ , &
\end{flalign*}
where the expectation is another Gaussian integral that can be computed via the unscented approximation as described in section \ref{sec:app_UT}. 
\vspace{-0.25cm}
\begin{figure}[h]
  \begin{center}
    \includegraphics[keepaspectratio=true]{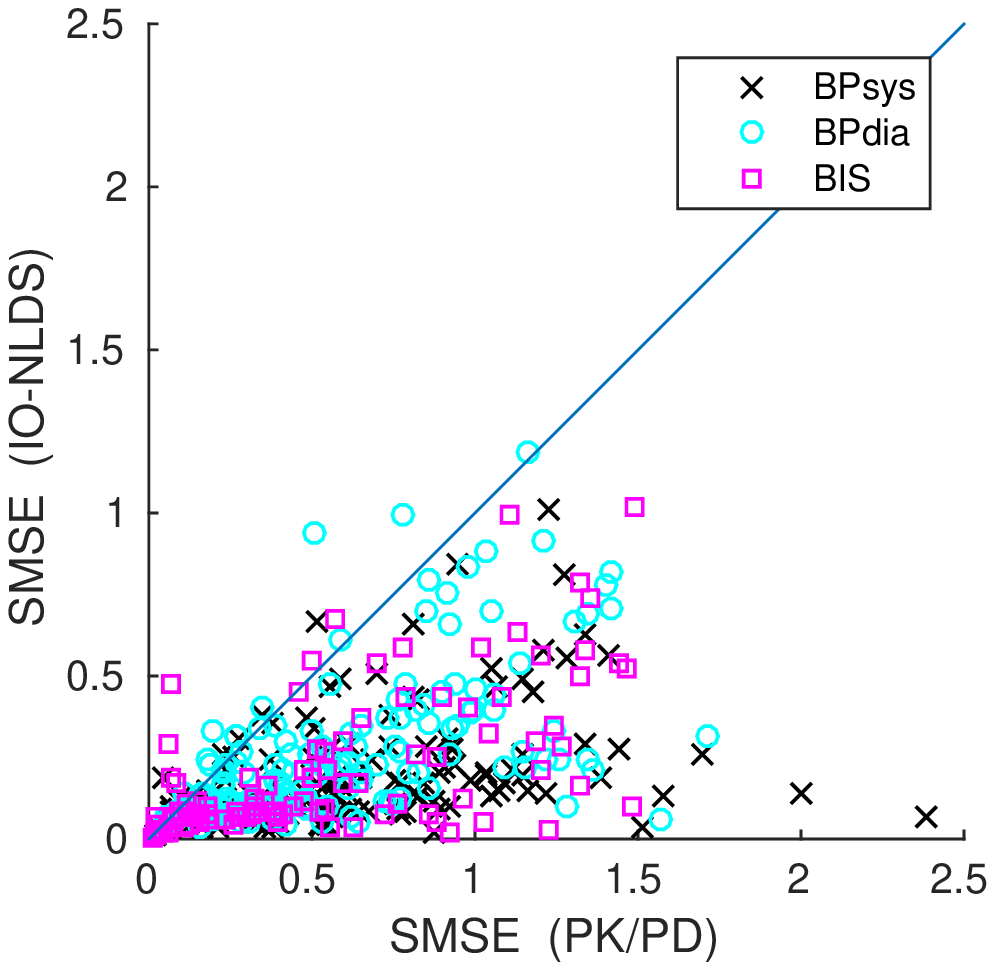}
  \end{center}
\caption*{Figure S1: SMSE results for all patients across all four prediction horizons and all three measured channels. Points below the 45 degree line correspond to lower errors for the IO-NLDS on the same task and points above that line correspond to lower errors for the PK/PD model. The diagonal line represents equal performance between the two models.}
\label{fig:smse_all}
\end{figure}
\end{document}